\documentclass{article}

\usepackage[final]{nips_2018}

\usepackage[utf8]{inputenc} 
\usepackage[T1]{fontenc}    
\usepackage{hyperref}       
\usepackage{url}            
\usepackage{booktabs}       
\usepackage{amsfonts}       
\usepackage{nicefrac}       
\usepackage{microtype}      
\usepackage{subfiles}
\usepackage{enumitem,kantlipsum}
\usepackage{graphicx}
\usepackage{caption}
\usepackage{subcaption}
\usepackage{tablefootnote}
\usepackage[bottom]{footmisc}
\graphicspath{ {./images/} }

\title{Using AI to Design Stone Jewelry}
\author{Khyatti Gupta, Sonam Damani, Kedhar Nath Narahari\\
\texttt{\{khgupt, sodamani, kedharn\}@microsoft.com}
}

\begin{document}

\maketitle
\begin{abstract}

Jewelry has been an integral part of human culture since ages. One of the most popular styles of jewelry is created by putting together precious and semi-precious stones in diverse patterns. While technology is finding its way in the production process of such jewelry, designing it remains a time-consuming and involved task. In this paper, we propose a unique approach using optimization methods coupled with machine learning techniques to generate novel stone jewelry designs at scale. Our evaluation shows that designs generated by our approach are highly likeable and visually appealing.

\end{abstract}

\section{Introduction}
\vspace{-0.3cm}

Jewelry has always been a significant part of human lifestyle and culture. Several styles of jewelry exist, ranging from bead work to stone setting, each characterized by the variety of materials and techniques used to make them. 
In recent years, jewelry manufacturing has made significant progress with intelligent machinery[8][6]. In the field of jewelry designing, AI is being progressively infused using algorithms based on plant modelling[9][10] and Voronoi diagrams[2] to create rubber, steel and brass jewelry.



Stone jewelry[1] is a conventional form of jewelry that involves setting precious and semi-precious stones, like Amethyst and Garnet, in uniquely crafted patterns.
Setting a combination of stones to look aesthetically appealing makes creation of stone jewelry a time-consuming and involved process. The system, described in this paper, models this problem as filling up of a given space by placing different shapes and sizes together. In this respect, research has been done to place non-overlapping geometric shapes together[11][7][4][3].  
However, such methods tend to optimize on dense packing rather than visual appeal, and hence are inadequate for creating coherent and aesthetic jewelry designs. 
In our work, we leverage jewelry design principles[5] to dynamically choose the most optimal stone placements for generating visually appealing designs. This is followed by a machine learning model to prune the unappealing designs. To the best of our knowledge, ours is the first work of applying AI to create designs for stone jewelry.

\section{Our Approach}

\vspace{-0.3cm}
The following section briefly describes our two-fold approach. For the purpose of this paper, we use 105 kinds of stones, each in 7 distinct shapes and 20 distinct sizes. 

\subsection{Generation}
\label{sec:generation}

\begin{table*}[!t]
\centering
\vspace{0cm}
\caption{\small Feature set used for pruning model of jewelry designs}
\label{feature-set}
\vspace{0cm}
\resizebox{\textwidth}{!}{%
  \begin{tabular}{p{3.8cm}p{13.3cm}}
    \toprule	
  \textbf{Features} & \textbf{Description} \\
    \toprule
    Balance & Distance between centroid of the jewelry shape and mean of centroids of all stones.\\
    Emphasis & Difference between area occupied by the biggest stone and mean area occupied by the remaining stones, multiplied by the standard deviation of area occupied by the remaining stones. \\
    Harmony of Shape & Standard deviation of shape vector of the jewelry, where shape vector is the set of counts of occurence for each available shape.\\
    Harmony of Orientation & Standard deviation of set of orientations of all stones.\\
    Proportion & Standard deviation of set of area occupied by all stones.\\
    Unity & Standard deviation of mean-adjacent-space for all stones, where mean-adjacent-space for a stone is the mean of space from all of its  neighbours.\\
    \bottomrule
  \end{tabular}}
  \vspace{0cm}
\end{table*}
\vspace{0cm}

We model the problem of jewelry designing as a packing problem[11], with jewelry shape as the container, stones shapes as objects, and jewelry design principles as the optimization function. 


The stones (objects) are placed in jewelry shape (container) one after the other. For placement of each stone:

\begin{enumerate}
    \item We find a candidate set of stones, whose placement shall not potentially create negative empty space at the end of placement of further stones, using dynamic programming.
    \item We rank the candidates using a set of rules based on their size, shape, orientation, and co-ordinates of the current position. The rules are optimized to maximize harmony, proportion, unity and balance defined in Table~\ref{feature-set}. 
    For instance, stone with size closer to the mean size of the stones that have been placed so far is ranked higher than the one which are farther. Similarly, a stone is ranked higher if its orientation is radially outward or inward to that of the stone placed opposite to it. 
    \item The candidate stone that is ranked the highest is chosen for the placement. 
\end{enumerate}

\subsection{Pruning}
\vspace{-0.2cm}
In order to discard the designs which are not as appealing, we use a pruning model. We create a dataset of 1200 designs generated using the approach described in Section 2.1, and get it annotated by 3 judges on the basis of their liking for each design. We use this dataset to train Gradient Boosted Trees, with features derived using rules adhering to jewelry design principles, as described in Table~\ref{feature-set}.


For the resultant designs, each stone is surrounded by a silver bezel and filled with textures of real stones. The whole process is visually depicted in Figure 1a.


\section{Experimetal Set-Up And Results}
\vspace{-0.3cm}
\begin{figure}[!t]
  \centering
  \vspace{0cm}
  \begin{subfigure}[b]{0.65\textwidth}
      \includegraphics[width=1\textwidth]{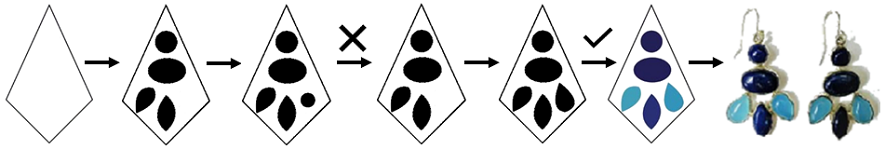}
      \vspace{-0.4cm}
      \caption{}
      \vspace{0cm}
      \label{fig:flow}
  \end{subfigure}
  \hspace{0.8cm}
  \begin{subfigure}[b]{0.11\textwidth}
      \includegraphics[width=1\textwidth]{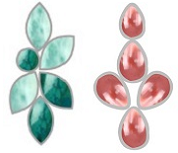}
      \vspace{-0.4cm}
      \caption{}
      \vspace{0cm}
       \label{fig:good}
  \end{subfigure}
  \hspace{0.8cm}
  \begin{subfigure}[b]{0.045\textwidth}
      \includegraphics[width=1\textwidth]{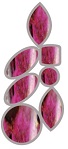}
      \vspace{-0.4cm}
      \caption{}
      \vspace{0cm}
       \label{fig:bad}
  \end{subfigure}
  \vspace{-0.1cm}
  \caption{\small{Designs generated: (a) Process; (b) Good; (c) Bad}}
  \vspace{-0.6cm}
\end{figure}


A random sample of 100 designs generated using our approach is used for evaluation. In order to account for diverse perceptions of beauty, each design is judged by 15 annotators. Also, since the personal bias inherently prevents consensual vote per design, the quality is measured for the entire set as a whole.

It is observed that at least 53\% of the designs created by our generative approach are liked by 53\% of the annotators. Applying pruning model to these designs further increases both the numbers by another 7\%. Many of the designs generated using our approach are aesthetically appealing, as shown in Figure~\ref{fig:good}. However, due to factors like disproportionate placements, lack of harmonic orientations, etc. some designs are not as admirable, as seen in Figure~\ref{fig:bad}. 




\section{Conclusion}
\vspace{-0.3cm}



In this paper, we develop a novel approach of creating stone jewelry designs at scale. We leverage jewelry design principles and dynamic programming for generating patterns using stones, followed by a pruning model to filter out the bad ones. Our approach generates diverse and visually aesthetic designs that are likeable and readily consumable for production. The last stage in Figure~\ref{fig:flow} shows a pair of ear rings made by craftsmen in the town of Jaipur, India using one of our designs. In our future work, we plan to make quality improvements by including additional design principles and extending our current work to other forms of jewelry.

\section*{References}

\small   
[1] Semi-Precious Stone \& Gemstone Jewelry - The Exclusive Indian Art. https://www.culturalindia. net/jewellery/types/stone-jewelry.html. [Online; accessed 26-October-2018]. 

[2] M. Balzer. Capacity-constrained voronoi diagrams in continuous spaces. In {\it 2009 Sixth International Symposium on Voronoi Diagrams} , pages 79–88, jun 2009. 

[3] T.C. Martins and M.S.G. Tsuzuki. Simulated annealing applied to the rotational polygon packing. {\it IFAC Proceedings Volumes} , 39(3):475–480, 2006. 

[4] K.A. Dowsland, Subodh Vaid, and W. B. Dowsland. An algorithm for polygon placement using a bottom-left strategy. {\it European Journal of Operational Research} , 141(2):371–381, 2002. 

[5] M.P. Galli, N. Giambelli, and F.Li. {\it The Art of Jewelry Design: Principles of Design, Rings and Earrings} . Schiffer Pub., 1994. 

[6] V.G. Gokhare, D.N. Raut, and D.K. Shinde. A Review paper on 3D-printing aspects and various processes used in the 3d-printing. {\it International Journal of Engineering Research \& Technology (IJERT)} , 6:953–958, 2017. 

[7] S. Jakobs. On genetic algorithms for the packing of polygons. {\it European journal of operational research} , 88(1):165–181, 1996. 

[8] L.C. Molinari, M.C. Megazzini, and E. Bemporad. The role of CAD/CAM in the modern jewellery business. {\it Gold Technology} , 23:3–7, 1998. 

[9] P. Prusinkiewicz, M. Hammel, J. Hanan, and R. Mech. L-systems: from the theory to visual models of plants. In {\it Proceedings of the 2nd CSIRO Symposium on Computational Challenges in Life Sciences} , volume 3, pages 1–32. CSIRO Publishing Melbourne, 1996. 

[10] A. Runions, B. Lane, and P. Prusinkiewicz. Modeling Trees with a Space Colonization Algorithm. {\it NPH} , 7:63–70, 2007. 

[11] D.R. Ulm and J.W. Baker. Solving a 2D Knapsack Problem on an Associative Computer Augmented with a Linear Network. In {\it PDPTA} , pages 29–32, 1996


\end{document}